\documentclass[10pt,twocolumn]{article}

\usepackage[utf8]{inputenc}
\usepackage[T1]{fontenc}
\usepackage{times}
\usepackage{amsmath, amssymb, amsthm}
\usepackage{graphicx}
\usepackage{booktabs}
\usepackage{multirow}
\usepackage{xcolor}
\usepackage{hyperref}
\usepackage{url}
\usepackage{microtype}
\usepackage{caption}
\usepackage{subcaption}
\usepackage{enumitem}
\usepackage{algorithm}
\usepackage{algorithmic}
\usepackage{geometry}
\usepackage{soul}
\usepackage[normalem]{ulem}
\usepackage{float}
\usepackage{placeins}
\usepackage{tikz}
\usetikzlibrary{positioning, calc, arrows.meta}

\title{
    \textbf{NepScript Genesis: Neural Architecture Search for Handwritten Devanagari Digit Synthesis}\\
    \vspace{0.5em}
}

\author{
    Mausam Gurung \\
    \texttt{mausaam.gurung593@gmail.com}
    \and
    Prabin Neupane \\
    \texttt{neupaneprabin2058@gmail.com}
    \and
    Sajjan Acharya \\
    \texttt{sajjanacharya11@gmail.com}
}

\date{\today}

\begin{document}
\maketitle

\vspace{1em}

\begin{abstract}

This paper introduces NepScript Genesis, a Neural Architecture Search (NAS) framework for automated Generative Adversarial Network (GAN) discovery, applied to conditional Devanagari handwritten digit synthesis. We compare five NAS strategies against a carefully constructed Deep Convolutional GAN (DCGAN) baseline ($FID=332.28$). Architecture selection utilizes a two-stage pipeline guided by a novel domain-aware evaluation metric (Enhanced Score). Results demonstrate that Adaptive Exploration achieves the optimal quality-efficiency trade-off, attaining an $FID$ of $79.12$—a $76.19\%$ improvement over the baseline—and the highest mode coverage among the NAS strategies ($Recall=0.531$) in under one GPU-hour. Furthermore, we demonstrate that incorporating script-specific structural heuristics into the search phase prevents early-stage mode collapse. In a downstream low-resource evaluation, augmenting 250 real
training samples per class with GAN-generated digits from the 
best NAS model improves CNN classification accuracy from 
$91.0\%$ to $96.5\%$ ($+5.5$ percentage points), demonstrating
that NAS-optimized synthesis produces digits of sufficient 
quality to benefit practical recognition pipelines when real 
data is scarce.


\vspace{0.5em}

\end{abstract}
 
\section{Introduction}
\label{sec:intro}

The synthesis of handwritten characters from low-resource scripts is an
important yet underexplored problem in generative modeling.
Devanagari~\cite{pant2012devanagari_multipurpose}, the script underlying
Nepali, Hindi, Sanskrit, and several other South Asian languages, presents
particular challenges for generative models: its characters feature complex
stroke topologies, conjunct consonant forms, vowel diacritics, and a
distinctive horizontal headline (\textit{shirorekha}) that must be faithfully
reproduced for generated samples to be legible. Automated synthesis of such
script is practically valuable for data augmentation in OCR
pipelines~\cite{guha2024devanagari_gan,mishra2021dcgan_devanagari},
educational tool development, and digital preservation of handwritten
documents in low-resource language settings~\cite{bhunia2019lowresource}.

Yet GAN training remains sensitive to architectural choices: network depth, normalization scheme, activation functions, latent space dimensionality, and regularization must all be selected by the practitioner through costly trial-and-error iteration. For a script as structurally demanding as Devanagari, poor architectural choices lead directly to mode collapse or failure to reproduce stroke topology.

Neural Architecture Search (NAS)~\cite{he2019automl} replaces this manual process with automated exploration guided by performance feedback. NAS has been extended to generative settings through AdversarialNAS~\cite{gao2019adversarialnas} and AutoGAN~\cite{gong2019autogan}, but has not been systematically applied to GAN architecture search for Devanagari or other Brahmic script synthesis tasks.
 
We address this gap with \textbf{NepScript Genesis}, a NAS framework that automates GAN architecture discovery for conditional Devanagari digit synthesis. We compare five search strategies against a manually designed DCGAN baseline and show that automated search consistently recovers better architectures within a practical single-GPU budget.

We make the following contributions:

\begin{enumerate}[leftmargin=*, itemsep=3pt]
    \item We design and evaluate a \textbf{manual DCGAN
    baseline}~\cite{radford2015dcgan} for Devanagari digit synthesis,
    establishing reference FID, IS, Precision, and Recall scores against
    which all NAS-discovered architectures are compared.
    {Manual DCGAN: FID $= 332.28$ | IS $= 2.00$ | P $= 0.693$ |
    R $= 0.552$}

    \item We implement and compare \textbf{five NAS strategies} --- Random
    Search~\cite{li2020randomnas}, Progressive Search, Adaptive
    Exploration, Multi-fidelity
    Search~\cite{li2017hyperband,jamieson2016successivehalving}, and
    AdversarialNAS~\cite{gao2019adversarialnas} --- under a fixed evaluation
    budget, demonstrating that intelligent search strategies consistently
    improve upon the manual baseline.
    
    \item We show that \textbf{Adaptive Exploration achieves the best
    quality-efficiency trade-off}, surpassing the manual DCGAN baseline
    on FID by 76.2\% (332.28 $\to$ 79.12) while requiring only 56.56
    minutes of search time --- less than half the cost of AdversarialNAS
    (124.50 min).

    \item We release an \textbf{open-source reproducible
    framework}~\cite{pineau2021reproducibility} for NAS experimentation on
    Devanagari and other Brahmic scripts, publicly available at
    \url{https://github.com/M9star/NepScript-Genesis-Framework}.
\end{enumerate}

\section{Related Work}
\label{sec:related}

\subsection{Generative Adversarial Networks}
\label{sec:related:gans}
Architectural advances in GAN --- including DCGAN~\cite{radford2015dcgan}, Progressive GAN~\cite{karras2018progressive}, StyleGAN~\cite{karras2019stylegan}, and BigGAN~\cite{brock2019biggan} --- have dramatically improved image quality and stability.
 
Training stability has been improved through Wasserstein
divergence~\cite{arjovsky2017wgan}, gradient penalty
regularization~\cite{gulrajani2017gradpenalty}, and spectral
normalization~\cite{miyato2018spectralnorm}. Normalization inside the network
is commonly handled with batch normalization~\cite{ioffe2015batchnorm} or
instance normalization~\cite{ulyanov2016instancenorm}. The FID
score~\cite{heusel2017fid} has become the standard evaluation metric for
generative models, as it correlates well with perceptual quality and captures
both fidelity and diversity. Conditional image synthesis --- enabling
class-conditioned generation --- was formalized by Mirza and
Osindero~\cite{mirza2014cgan}.

\subsection{Neural Architecture Search}
\label{sec:related:nas}

NAS methods can be categorized by their search strategy and performance
estimation strategy~\cite{he2019automl}.

\textbf{Search strategies:}
\begin{itemize}[leftmargin=*, itemsep=2pt]
    \item \textit{Random search:} Uniform sampling from the configuration
    space; a surprisingly competitive baseline that is often underestimated
    in high-dimensional spaces~\cite{li2020randomnas}. It establishes a
    lower bound on the difficulty of a search problem: if random search
    performs well, the landscape is relatively uniform; if it performs
    poorly, structured search strategies provide clear value.
    \item \textit{Reinforcement learning:} Uses a controller trained via
    policy gradients to propose architectures, rewarding configurations
    that achieve high validation performance. NASNet~\cite{zoph2017nasnet}
    and ENAS~\cite{pham2018enas} demonstrate strong results but require
    thousands of GPU-hours for full search.
    \item \textit{Evolutionary:} Maintains a population of architectures
    and iteratively mutates and selects the fittest candidates.
    AmoebaNet~\cite{real2019amoebanet} shows evolutionary search can match
    or exceed RL-based methods at comparable cost.
    \item \textit{Gradient-based:} Relaxes discrete architecture choices
    to continuous parameters, enabling joint optimization of architecture
    and weights via gradient descent. DARTS~\cite{liu2019darts} and
    SNAS~\cite{xie2019snas} reduce search cost to tens of GPU-hours but
    can overfit the supernet.
    \item \textit{Multi-fidelity:} Evaluates candidates at reduced
    fidelity (fewer epochs, smaller datasets) and promotes only promising
    ones to full evaluation. Successive Halving~\cite{jamieson2016successivehalving}
    and Hyperband~\cite{li2017hyperband} formalize this as a bandit
    problem, providing theoretical guarantees on sample efficiency.
        
\end{itemize}

\textbf{NAS for generative models:}
AdversarialNAS~\cite{gao2019adversarialnas} introduces gradient-based joint optimization of $G$ and $D$ architectures, directly targeting GAN training stability. AutoGAN~\cite{gong2019autogan} applies an RL-based controller to progressively search generator architectures at multiple resolutions. A recent survey~\cite{alotaibi2025nasgan_survey} reviews the broader field of automated GAN design, confirming that systematic NAS for low-resource script synthesis remains an open problem.

\textbf{Efficiency of NAS:} Gradient-based NAS (DARTS~\cite{liu2019darts}) can reduce architecture search from thousands of GPU-hours (RL-based methods) to tens of GPU-hours by sharing weights across the supernet. Multi-fidelity methods~\cite{li2017hyperband} achieve further savings by early stopping of poor candidates.
\subsection{Handwritten Character Synthesis and Devanagari}
\label{sec:related:devanagari}

Prior work on Devanagari handwriting has largely focused on recognition rather
than synthesis. The Devanagari Handwritten Character Dataset
(DHCD)~\cite{acharya2015dhcd,acharya2015dhcd_dataset} established a standard
benchmark for recognition tasks. Synthesis efforts include DCGAN-based
augmentation~\cite{mishra2021dcgan_devanagari}, conditional GAN
generation~\cite{giri2021cgan_devanagari}, and GAN-based synthetic data for
improving recognizer accuracy~\cite{guha2024devanagari_gan}. Low-resource
script generation more broadly has been studied through adversarial
learning~\cite{bhunia2019lowresource}. A multi-purpose Devanagari numeral
dataset is provided by Pant et al.~\cite{pant2012devanagari_multipurpose}.

To our knowledge, NepScript Genesis is the first work to apply systematic NAS
to GAN architecture search for Devanagari digit synthesis, bridging the gap
between automated architecture search and low-resource script generation.

\section{Methodology}
\label{sec:method}

\subsection{Problem Formulation}
\label{sec:method:formulation}

We frame the task as conditional image synthesis. Given a dataset
$\mathcal{D} = \{(x_i, y_i)\}_{i=1}^N$ of handwritten Devanagari digit images
with class labels $y_i \in \{0,1,\ldots,9\}$, we seek to learn a generator
$G_\theta: \mathcal{Z} \times \mathcal{Y} \to \mathcal{X}$ such that the
distribution of generated samples $p_g = G_\theta(z, y),\ z \sim p(z),\
y \sim p(y)$ approximates the real data distribution $p_{\text{data}}$
conditioned on digit class~\cite{mirza2014cgan}.

Generator and discriminator engage in the standard conditional adversarial
objective~\cite{goodfellow2014gan,mirza2014cgan}:

\begin{equation}
    \min_G \max_D\ \mathbb{E}_{x \sim p_{\text{data}}}[\log D(x)] +
    \mathbb{E}_{z \sim p(z)}[\log(1 - D(G(z)))]
    \label{eq:gan_objective}
\end{equation}

Rather than manually specifying $G$ and $D$, we automate architecture
selection via NAS over a predefined search space $\Lambda$. The NAS problem
is:
\begin{equation}
    \lambda^* = \arg\min_{\lambda \in \Lambda}\
    \mathcal{L}_{\text{eval}}(G_\lambda, D_\lambda, \mathcal{D})
    \label{eq:nas_objective}
\end{equation}
where $\mathcal{L}_{\text{eval}}$ is the FID score~\cite{heusel2017fid}
(lower is better), used as the primary criterion for architecture ranking.

\subsection{Manual DCGAN Baseline}
\label{sec:method:dcgan}

To establish a reference point for NAS evaluation, we design a
DCGAN~\cite{radford2015dcgan} baseline following established best practices.
The manual baseline architecture is:

\begin{itemize}[leftmargin=*, itemsep=2pt]
    \item \textbf{Generator:} 4 transposed convolutional layers, latent
    dim $= 100$, BatchNorm~\cite{ioffe2015batchnorm} after each layer except
    output, ReLU activations, Tanh output.
    \item \textbf{Discriminator:} 4 convolutional layers, LeakyReLU,
    BatchNorm, Sigmoid output.
    \item \textbf{Training:} Adam optimizer, lr $= 0.0002$,
    $\beta_1 = 0.5$, batch size $= 64$, discriminator updated every 2
    batches, $N = 500$ epochs.
\end{itemize}

\begin{figure}[h!]
\centering
\resizebox{\columnwidth}{!}{%
\begin{tikzpicture}[
    node distance=0.5cm and 0.6cm,
    block/.style={rectangle, draw, minimum width=1.8cm,
                  minimum height=1.0cm, align=center, font=\small},
    arrow/.style={->, >=stealth, thick}
]

\node[block] (z)  {$z \sim \mathcal{N}(0,I)$\\$d_z = 100$};
\node[block, right=of z]  (g1) {ConvT\\$100{\to}512$\\BN, ReLU};
\node[block, right=of g1] (g2) {ConvT\\$512{\to}256$\\BN, ReLU};
\node[block, right=of g2] (g3) {ConvT\\$256{\to}128$\\BN, ReLU};
\node[block, right=of g3] (g4) {ConvT\\$128{\to}1$\\Tanh};
\node[block, right=of g4] (fakeimg) {Fake Image\\$1{\times}32{\times}32$};

\draw[arrow] (z)   -- (g1);
\draw[arrow] (g1)  -- (g2);
\draw[arrow] (g2)  -- (g3);
\draw[arrow] (g3)  -- (g4);
\draw[arrow] (g4)  -- (fakeimg);

\draw[dashed] ($(z.north west)+(-0.2,0.4)$)
    rectangle ($(g4.south east)+(0.2,-0.4)$);
\node[above] at ($(z.north)!0.5!(g4.north)+(0,0.4)$)
    {\textbf{Generator $G$}};

\node[block, below=2.0cm of z] (realimg) {Real Image\\$1{\times}32{\times}32$};
\node[block, right=of realimg] (d1) {Conv\\$1{\to}64$\\LReLU};
\node[block, right=of d1]      (d2) {Conv\\$64{\to}128$\\BN, LReLU};
\node[block, right=of d2]      (d3) {Conv\\$128{\to}256$\\BN, LReLU};
\node[block, right=of d3]      (d4) {Conv\\$256{\to}1$\\Sigmoid};
\node[block, right=of d4]      (out){Real/Fake\\Score};

\draw[arrow] (realimg) -- (d1);
\draw[arrow] (d1) -- (d2);
\draw[arrow] (d2) -- (d3);
\draw[arrow] (d3) -- (d4);
\draw[arrow] (d4) -- (out);

\draw[dashed] ($(d1.north west)+(-0.2,0.4)$)
    rectangle ($(d4.south east)+(0.2,-0.4)$);
\node[above] at ($(d1.north)!0.5!(d4.north)+(0,0.4)$)
    {\textbf{Discriminator $D$}};

\draw[arrow] (fakeimg.south) -- ++(0,-0.4) -| (d1.north);

\draw[arrow, dashed] (out.east) -- ++(0.4,0)
    -- ++(0,3.5) -- (fakeimg.east)
    node[midway, right]{\small Gradient};

\end{tikzpicture}}
\caption{Manual DCGAN baseline: Generator $G$ (top)
and Discriminator $D$ (bottom)~\cite{radford2015dcgan}.}
\label{fig:dcgan_baseline}
\end{figure}
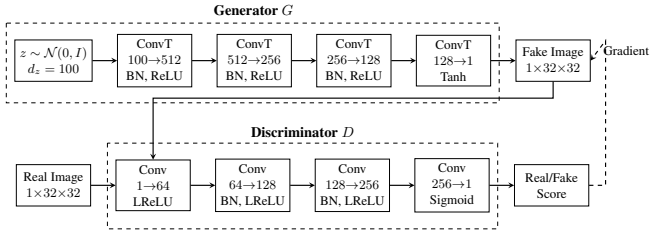
\FloatBarrier

The manual DCGAN baseline achieves:
\begin{center}
\begin{tabular}{lcccc}
\toprule
\textbf{Model} & \textbf{FID} $\downarrow$ & \textbf{IS} $\uparrow$ &
\textbf{Prec.} $\uparrow$ & \textbf{Recall} $\uparrow$ \\
\midrule
Manual DCGAN & 332.28 & 2.00 & 0.693 & 0.552 \\
\bottomrule
\end{tabular}
\end{center}

This is what a practitioner reaches by following standard DCGAN conventions
without any architecture search, and it sets the floor that NAS strategies
must clear to be worthwhile.

\subsection{Dataset}
\label{sec:method:dataset}

\begin{figure}[htbp]
    \centering
    \includegraphics[width=0.85\columnwidth, height=7cm, keepaspectratio]{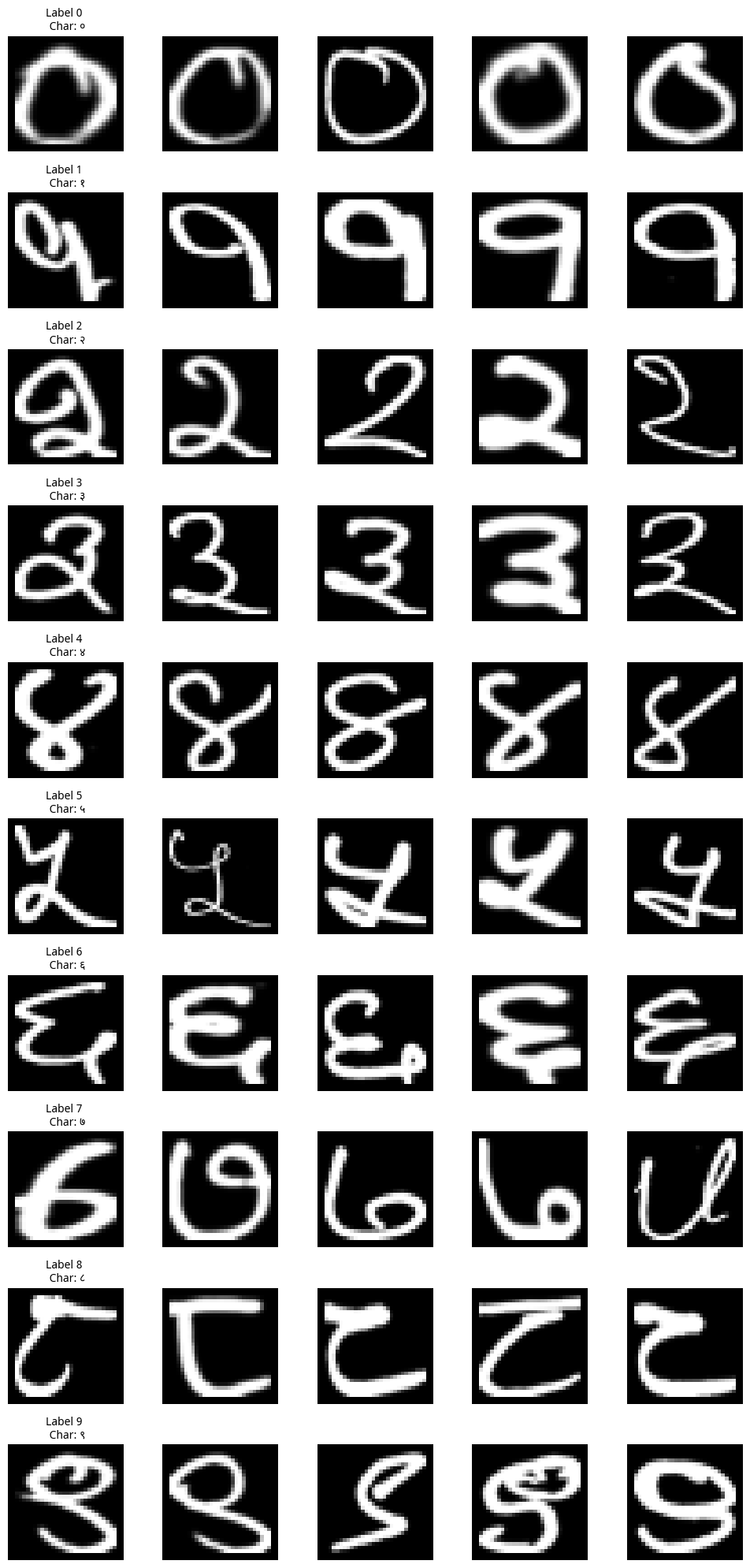}
    \caption{Sample grid showing 5 examples per digit class ($10 \times 5 = 50$ images).}
    \label{fig:dataset_samples}
\end{figure}

We use the Devanagari Handwritten Character
Dataset~\cite{acharya2015dhcd,acharya2015dhcd_dataset}, comprising
20{,}000 handwritten Devanagari digit samples spanning ten digit classes
(0--9 in Devanagari script), with 2{,}000 samples per class. Sample
images for each class are shown in Figure~\ref{fig:dataset_samples}.

Images are resized to $32 \times 32$ pixels and pixel values normalized to
$[-1, 1]$. The dataset is class-balanced.

\subsection{NAS Search Space}
\label{sec:method:searchspace}

The search space $\Lambda$ is defined over five architectural dimensions,
applied independently to both the Generator $G$ and Discriminator $D$:

\begin{table*}[t]
\centering
\caption{NAS Search Space Definition. Each dimension is searched
independently for $G$ and $D$, yielding a joint search space
substantially larger than the per-network count.}
\label{tab:searchspace}
\begin{tabular}{llll}
\toprule
\textbf{Dimension} & \textbf{Choices} & \textbf{Applied to} & \textbf{Motivation} \\
\midrule
Latent dim ($d_z$)   & \{64, 100, 128, 200\} & $G$ only
    & Controls generator capacity for ten-class conditional synthesis \\
Normalization        & \{BatchNorm, InstanceNorm\} & $G$ and $D$
    & \cite{ioffe2015batchnorm,ulyanov2016instancenorm} \\
Activation           & \{ReLU, LeakyReLU\} & $G$ and $D$
    & \cite{radford2015dcgan} \\
Dropout rate ($G$)   & \{0.0, 0.1, 0.2, 0.3\} & $G$ only
    & \cite{srivastava2014dropout} \\
Dropout rate ($D$)   & \{0.0, 0.2, 0.3, 0.5\} & $D$ only
    & \cite{srivastava2014dropout} \\
Residual connections & \{False, True\} & $G$ and $D$
    & \cite{he2016resnet} \\
\bottomrule
\end{tabular}
\end{table*}

The generator-side search space contains $4 \times 2 \times 2 \times 4
\times 2 = 128$ unique configurations. Since normalization, activation,
dropout, and residual connections are searched independently for $G$ and
$D$, the joint generator--discriminator space is $128 \times 2 \times 2
\times 4 \times 2 = 4{,}096$ unique paired configurations. This is far too
large to evaluate exhaustively, which is precisely why we turn to NAS rather
than manual search.

The five dimensions were selected to cover the primary sources of
architectural variation in DCGAN-style networks~\cite{radford2015dcgan}
while keeping the search tractable for a single-GPU setting. Deeper
network variants, attention mechanisms, and alternative loss functions
represent natural extensions of this search space and are left for
future work.

All configurations are trained with fixed hyperparameters: learning rate
$= 0.0002$, batch size $= 64$, label smoothing $=
0.1$~\cite{szegedy2016labelsmoothing}, Adam optimizer with $\beta_1 =
0.5$, $\beta_2 = 0.999$, consistent with the manual DCGAN baseline
(Section~\ref{sec:method:dcgan}).

\subsection{NAS Algorithms}
\label{sec:method:nasalgos}

We implement and evaluate five NAS strategies. Each operates over the
128-configuration search space $\Lambda$ defined in
Table~\ref{tab:searchspace}, using the Enhanced Score
(Equation~\ref{eq:enhanced_score}) after 30-epoch training as the
evaluation criterion.

\subsubsection{Random Search}

Uniformly samples from $\Lambda$ without learning between evaluations,
following Li and Talwalkar~\cite{li2020randomnas}. Serves as the primary
NAS baseline requiring no surrogate model.

\begin{algorithm}[H]
\caption{Random Search}
\begin{algorithmic}[1]
\STATE Initialize: results $\leftarrow$ [ ], best\_score $\leftarrow$ 0,
       tested $\leftarrow \emptyset$
\FOR{$i = 1$ \TO $N$}
    \STATE $\lambda \leftarrow$ \textsc{SampleRandom}($\Lambda \setminus$
           tested)
    \STATE score $\leftarrow$ \textsc{TrainEvaluate}($\lambda$, 30 epochs)
    \STATE Update best if score $>$ best\_score
\ENDFOR
\RETURN best architecture
\end{algorithmic}
\end{algorithm}

\subsubsection{Progressive Search}

Searches in three phases of increasing complexity, with $[4, 5, 6]$
architectures per phase (15 total). Phase 1 uses conservative search
spaces (small latent dims, no residual); Phase 3 uses aggressive spaces
(large latent dims, residual enabled)~\cite{karras2018progressive}.

\begin{algorithm}[H]
\caption{Progressive Search}
\begin{algorithmic}[1]
\STATE phases $\leftarrow$ [conservative, moderate, aggressive],
       targets $\leftarrow$ [4, 5, 6]
\FOR{phase $p = 1, 2, 3$}
    \STATE Set search space $\leftarrow \Lambda_p$
    \FOR{$i = 1$ \TO targets[$p$]}
        \STATE $\lambda \leftarrow$ \textsc{SampleRandom}($\Lambda_p$)
        \STATE score $\leftarrow$ \textsc{TrainEvaluate}($\lambda$,
               30 epochs)
        \STATE Update best if score $>$ best\_score
    \ENDFOR
\ENDFOR
\RETURN best architecture
\end{algorithmic}
\end{algorithm}

\subsubsection{Adaptive Exploration}

Maintains a history of successful configurations (score $\geq 0.5$) and
uses this history to bias sampling --- with probability $0.7$ mutating a
successful configuration, otherwise exploring randomly. Conceptually related to exploitation-biased evolutionary search.

\begin{algorithm}[H]
\caption{Adaptive Exploration}
\begin{algorithmic}[1]
\STATE Initialize: successful\_configs $\leftarrow$ [ ],
       iteration $\leftarrow$ 0
\WHILE{iteration $< N$}
    \IF{$|$successful\_configs$| < 3$}
        \STATE $\lambda \leftarrow$ \textsc{SampleRandom}($\Lambda$)
    \ELSIF{random() $< 0.7$}
        \STATE $c \leftarrow$ \textsc{RandomChoice}(successful\_configs)
        \STATE $\lambda \leftarrow$ \textsc{Mutate}($c$)
    \ELSE
        \STATE $\lambda \leftarrow$ \textsc{SampleRandom}($\Lambda$)
    \ENDIF
    \STATE score $\leftarrow$ \textsc{TrainEvaluate}($\lambda$, 30 epochs)
    \IF{score $\geq 0.5$}
        \STATE successful\_configs.append($\lambda$); keep last 10
    \ENDIF
    \STATE Update best; iteration $\leftarrow$ iteration $+ 1$
\ENDWHILE
\RETURN best architecture
\end{algorithmic}
\end{algorithm}

\subsubsection{Multi-fidelity Search}

Evaluates each candidate at three progressive fidelity stages, eliminating
poor performers early. Related to Successive
Halving~\cite{jamieson2016successivehalving} and
Hyperband~\cite{li2017hyperband}.

\begin{algorithm}[H]
\caption{Multi-fidelity Search}
\begin{algorithmic}[1]
\STATE Initialize: evaluation\_history $\leftarrow$ [ ]
\WHILE{architecture\_count $< N$}
    \STATE $\lambda \leftarrow$ \textsc{SampleRandom}($\Lambda$)
    \STATE score$_1$ $\leftarrow$ \textsc{TrainEvaluate}($\lambda$,
           10 epochs)
    \IF{score$_1 < 0.5$} \STATE \textbf{skip} \ENDIF
    \STATE score$_2$ $\leftarrow$ \textsc{TrainEvaluate}($\lambda$,
           20 epochs)
    \IF{score$_2 < 0.7$} \STATE record score$_2$; \textbf{continue}
    \ENDIF
    \STATE score$_3$ $\leftarrow$ \textsc{TrainEvaluate}($\lambda$,
           30 epochs)
    \STATE Update best if score$_3 >$ best\_score
\ENDWHILE
\RETURN best architecture
\end{algorithmic}
\end{algorithm}

\subsubsection{Adversarial Gradient-based Search (AdversarialNAS)}
\label{sec:method:adversarialnas}

Following~\cite{gao2019adversarialnas}, architecture parameters are
relaxed to continuous values and jointly optimized with model weights
via gradient descent, similar to DARTS~\cite{liu2019darts}. Generator
and discriminator architectures are represented as weighted mixtures
over candidate operations:

\begin{equation}
    \bar{o}^{(i,j)} = \sum_{o \in \mathcal{O}}
    \frac{\exp(\alpha_o^{(i,j)})}{\sum_{o'}\exp(\alpha_{o'}^{(i,j)})}
    \cdot o(x)
    \label{eq:darts_mixing}
\end{equation}

where $\alpha^{(i,j)}$ are learnable architecture parameters and
Gumbel-Softmax~\cite{jang2017gumbel} enables gradient flow through
discrete selections. Following the bilevel scheme of
DARTS~\cite{liu2019darts}, we hold out a validation split and alternate two
updates each epoch: network weights $w_G, w_D$ are updated on a
\emph{training} batch via the adversarial loss, while the architecture
parameters $\alpha_G, \alpha_D$ are updated on a \emph{validation} batch, so
that $\alpha$ is optimized for validation performance rather than training
fit. After search, the discrete architecture is derived by taking the
$\arg\max$ over each $\alpha^{(i,j)}$.

\begin{algorithm}[H]
\caption{Adversarial Gradient-based Search}
\begin{algorithmic}[1]
\STATE Initialize weights $w$ and architecture parameters $\alpha$
\WHILE{epoch $< N$}
    \STATE Sample training batch $B_{\text{tr}}$, validation batch
           $B_{\text{val}}$
    \STATE $w_G, w_D \leftarrow$ update on $B_{\text{tr}}$ by adversarial
           loss \COMMENT{weights}
    \STATE $\alpha_G, \alpha_D \leftarrow$ update on $B_{\text{val}}$ by
           validation loss \COMMENT{architecture}
    \STATE epoch $\leftarrow$ epoch $+ 1$
\ENDWHILE
\RETURN architecture from $\arg\max_o \alpha^{(i,j)}_o$
\end{algorithmic}
\end{algorithm}

\subsection{Evaluation Metrics}
\label{sec:method:metrics}

We employ a \textbf{two-stage evaluation pipeline} to separate architecture
search from final performance reporting.

\textbf{Stage 1 --- Architecture Filtering (30 epochs):}
Each candidate architecture is trained for 30 epochs and ranked using a
composite \textit{Enhanced Score} — a weighted combination of training
health, output diversity, structural consistency, convergence efficiency,
and script-specific visual quality:

{\small 

\begin{multline}
    \textbf{Enhanced Score} = 0.30 \times \text{Base} + 0.20 \times \text{Diversity} \\
    + 0.20 \times \text{Consistency} + 0.15 \times \text{Efficiency} \\
    + 0.15 \times Q_{\text{Nepali}}
    \label{eq:enhanced_score}
\end{multline}
}

\textbf{Base Score} measures fundamental training health:
{\small
\begin{equation}
    \textbf{Base} = 0.4 \times \text{Stability} + 0.3 \times \text{Balance}
    + 0.3 \times \text{Quality}
    \label{eq:base_score}
\end{equation}
}

where Stability tracks training consistency over time:
{\small
\begin{equation}
    \textbf{Stability} = \frac{\text{Stable Epochs}}{\text{Reference Epochs (10)}}
\end{equation}}

Balance measures proximity of generator $G$ and discriminator $D$ losses
to the ideal value of $1.0$:
{\small
\begin{equation}
    \textbf{Balance} = \frac{\max(0,\, 1 - |G_{\ell} - 1|)
    + \max(0,\, 1 - |D_{\ell} - 1|)}{2}
\end{equation}}

and Quality assesses pixel integrity via intensity $\bar{I}_c$ and standard
deviation $\sigma_c$ across $N$ classes:
{\small
\begin{equation}
    \textbf{Quality} = \frac{1}{N} \sum_{c=1}^{N}
    \frac{\max(0,\, 1 - |\bar{I}_c - 0.5|) + \min(1,\, 2\sigma_c)}{2}
\end{equation}}

\textbf{Diversity} penalizes mode collapse via average pairwise distance
between generated samples:
{\small
\begin{equation}
    \text{Diversity} = \min\!\left(1,\;
    \frac{\text{Avg. Pairwise Distance}}{2.0}\right)
    \label{eq:diversity}
\end{equation}}

\textbf{Consistency} measures label-conditional stability across $B$ batches
and $K$ classes:
{\small
\begin{equation}
    \text{Consistency} = \frac{1}{K}\sum_{k=1}^{K}
    \frac{1}{B}\sum_{b=1}^{B} \frac{1}{1 + \text{Var}(X_{k,b})}
\end{equation}}

Here $\text{Var}(X_{k,b})$ is the pixel variance among samples generated for
class $k$ in batch $b$, so Consistency rewards a label reliably producing a
coherent digit rather than collapsing to noise. It therefore pulls against
Diversity (Eq.~\ref{eq:diversity}), which rewards cross-sample variation;
the two carry equal weight ($0.20$ each) so the Enhanced Score favours
outputs that are recognizable per class yet varied across samples, rather
than maximizing either alone.

\textbf{Efficiency} rewards fast convergence and compact latent spaces:
{\small
\begin{equation}
    \text{Efficiency} = \frac{1}{2}\!\left[
    \frac{1}{1 + |G_{\ell} - D_{\ell}|}
    + \frac{1}{1 + \frac{d_z}{100}}\right]
\end{equation}}

where $d_z$ is the latent dimension and $G_\ell$, $D_\ell$ denote the final
generator and discriminator losses respectively. The compactness sub-term
$1/(1{+}d_z/100)$ is a soft preference, not a hard penalty: Efficiency
carries only 15\% of the Enhanced Score, so its effect is small (a shift
from $d_z{=}64$ to $d_z{=}128$ changes the total score by ${\sim}0.01$). In
practice it did not override the other terms --- Adaptive Exploration still
selected $d_z{=}128$ --- so the larger latent space was retained where it
improved quality.

\textbf{Nepali Quality} We introduce $Q_{Nepali}$ as a domain-aware heuristic to strictly penalize topological failures in early training. By explicitly rewarding stroke integrity ($S_{stroke}$) and variance-based sharpness ($S_{sharp}$), the search algorithm discards topologically unsound architectures prior to full convergence.

{\small
\begin{equation}
    Q_{\text{Nepali}} = \frac{1}{N}\sum_{i=1}^{N}
    S_{\text{sharp}}(x_i) \times S_{\text{stroke}}(x_i)
\end{equation}}

where sharpness $S_{\text{sharp}} = \min\!\left(1,\,
\frac{\text{Var}(\nabla^2 x_i)}{0.15}\right)$ uses the Laplacian variance,
and stroke integrity $S_{\text{stroke}}$ is based on the number of connected
components $n$: $1.0$ if $n{=}1$, $0.9$ if $n{=}2$, and
$\max(0,\, 1 - (n-2) \times 0.25)$ if $n{>}2$.

The architecture with the highest Enhanced Score per strategy is selected
and passed to Stage 2.

\textbf{Stage 2 --- Full Training and Final Evaluation (500 epochs):}
The selected architecture is retrained from scratch for 500 epochs and
evaluated using four standard metrics:

\textbf{FID}~\cite{heusel2017fid} (lower is better):
\begin{equation}
    \text{FID} = \|\mu_r - \mu_g\|^2 +
    \text{Tr}\!\left(\Sigma_r + \Sigma_g - 2\sqrt{\Sigma_r \Sigma_g}\right)
    \label{eq:fid}
\end{equation}

\textbf{Inception Score (IS)}~\cite{salimans2016is} (higher is better):
\begin{equation}
    \text{IS} = \exp\!\left(\mathbb{E}_{x \sim p_g}\!\left[
    D_{\text{KL}}\!\left(p(y|x) \,\|\, p(y)\right)\right]\right)
    \label{eq:is}
\end{equation}

\textbf{Precision \& Recall}~\cite{kynkaanniemi2019precision}:
\begin{align}
    \text{Precision} &= \frac{|\{x_g : x_g \in
    \text{supp}(p_{\text{real}})\}|}{|\{x_g\}|}
    \label{eq:precision} \\
    \text{Recall} &= \frac{|\{x_r : x_r \in
    \text{supp}(p_{\text{gen}})\}|}{|\{x_r\}|}
    \label{eq:recall}
\end{align}

Precision measures sample fidelity; Recall measures mode coverage. A model
with high Precision but low Recall is mode-collapsing; high Recall but low
Precision indicates diversity at the cost of fidelity. All Stage 2 FID
values are computed on 20{,}000 samples for
consistency~\cite{borji2019fidsensitivity}.

\textbf{Justification of 30-epoch proxy.}
The 30-epoch filtering stage follows the multi-fidelity
principle~\cite{li2017hyperband,jamieson2016successivehalving} that short
runs yield reliable \emph{relative} rankings before convergence, especially
when the criterion captures training dynamics rather than final output
quality~\cite{li2020randomnas}. The Enhanced Score is built for this
setting: measured at 30 epochs, its Stability, Balance, and diversity terms
indicate which architectures will train well over 500 epochs at a fraction
of the cost. We use it strictly as a \emph{within-strategy} criterion --- to
select the best candidate each strategy proposes, not to predict absolute
final FID across strategies. The best scores from all strategies fall in a
narrow band (0.976--0.989), yet within each strategy the Enhanced Score
spreads candidates widely (per-strategy ranges of 0.15--0.47). The proxy
therefore discriminates effectively among the architectures a strategy
proposes, even though the per-strategy winners cluster tightly: the FID gap
across strategies reflects the quality of candidates each \emph{proposes},
not differences in proxy score, and Adaptive Exploration's advantage is
proposing better candidates.

\subsection{CNN Classifier for Downstream Evaluation}
\label{sec:method:cnn}

To assess the practical utility of GAN-generated digits beyond distributional
metrics, we train a lightweight CNN classifier in a low-resource
setting and measure whether augmenting real training data with
GAN-generated samples improves recognition accuracy.

\textbf{Architecture.} The classifier is intentionally lightweight to ensure
the baseline has meaningful headroom for improvement, making the effect of
GAN augmentation clearly observable. It comprises a single convolutional
layer ($5\times5$ kernel, 8 filters) followed by ReLU activation and
$2\times2$ max pooling, yielding feature maps of size $14\times14$. The
flattened $8\times14\times14$ representation is projected through a 32-unit
hidden layer (dropout $p=0.6$, ReLU) to a 10-class output. The model
contains significantly fewer parameters than standard CNN classifiers,
making it sensitive to training data quantity and quality.

\textbf{Training protocol.} The model is trained with Adam ($\text{lr} =
0.001$, batch size $= 32$) for 30 epochs with no early stopping. Training
images use light augmentation (random rotation $\pm10^{\circ}$, random translation
$\pm10\%$); test images use only resize and normalize. Random seed is fixed
at 42 for reproducibility. The validation split (10\% of training data) is
used only for monitoring --- final evaluation is performed separately on the
held-out test set (3{,}000 samples, 300 per class).

\textbf{Evaluation conditions.} We simulate a single low-resource scenario
by limiting real training data to 250 samples per class (2{,}500
total), and evaluate two conditions on the same held-out test set
(3{,}000 samples, 300 per class):

\begin{enumerate}[leftmargin=*, itemsep=2pt]
    \item \textbf{Baseline:} CNN trained on 250 real samples per
    class only (2{,}500 total).
    \item \textbf{GAN-augmented:} CNN trained on 250 real samples
    per class augmented with approximately 1{,}000 GAN-generated
    samples per class (9{,}935 total) from the best NAS model ---
    Adaptive Exploration (Section~\ref{sec:results:arch_analysis}),
    yielding approximately 1{,}250 training samples per class.
\end{enumerate}

Both conditions use identical architecture, hyperparameters, and
training protocol. Results are reported in
Section~\ref{sec:results:downstream}.

\section{Experiments and Results}
\label{sec:results}


\subsection{Experimental Setup}
\label{sec:results:setup}

All experiments use the dataset of Section~\ref{sec:method:dataset}. The
four sampling strategies (Random, Progressive, Adaptive Exploration, and
Multi-fidelity) each evaluate $N = 15$ candidates by Enhanced Score
(Equation~\ref{eq:enhanced_score}) after short-epoch filtering, whereas
AdversarialNAS instead refines a single architecture over ${\sim}300$
gradient-based iterations. The best architecture per strategy is then
retrained from scratch for 500 epochs (Table~\ref{tab:main_results}).

All runs use an NVIDIA RTX 4050 GPU and PyTorch 2.0 with
Adam~\cite{radford2015dcgan} (learning rate $0.0002$, $\beta_1 = 0.5$,
$\beta_2 = 0.999$, batch size $64$). Owing to compute constraints we report
a single run per architecture; multi-seed estimation is left to future
work~\cite{pineau2021reproducibility}. Our aim is to quantify the benefit
of automated search over a manual design point, not to claim
state-of-the-art synthesis. For Multi-fidelity Search, candidates pass
through 10-, 20-, and 30-epoch stages with promotion thresholds of 0.5 and
0.7; those failing Stage 1 are discarded and those failing Stage 2 are
recorded but not promoted.

\subsection{NAS vs.\ Manual DCGAN Baseline}
\label{sec:results:vs_manual}

Table~\ref{tab:main_results} presents the main results comparing the
manual DCGAN baseline to all five NAS strategies.

\begin{table*}[t]
\centering
\caption{
    Performance comparison: Manual DCGAN baseline vs.\ NAS-discovered
    architectures on Devanagari digit synthesis. Best result per
    metric (across all rows, including the baseline) in \textbf{bold}.
    $\Delta$FID computed relative to manual
    DCGAN (332.28); all NAS strategies improve upon the baseline.
    $^\dagger$AdversarialNAS optimizes a single architecture over
    ${\sim}300$ gradient steps rather than evaluating 15 discrete
    candidates; search time is not directly comparable to
    sampling-based strategies.
}
\label{tab:main_results}
\begin{tabular}{lcccccc}
\toprule
\textbf{Method} & \textbf{FID} $\downarrow$ & \textbf{IS} $\uparrow$ &
\textbf{Precision} $\uparrow$ & \textbf{Recall} $\uparrow$ &
\textbf{Search Time (min)} & \textbf{$\Delta$FID vs.\ Manual} \\
\midrule
\multicolumn{7}{l}{\textit{Manual Baseline}} \\
Manual DCGAN          & 332.28 & 2.00 & 0.693 & \textbf{0.552} & -- & -- \\
\midrule
\multicolumn{7}{l}{\textit{NAS Strategies: sampling strategies use 15 configs $\times$ 30-epoch filtering $\to$ 500-epoch full training}} \\
Random Search~\cite{li2020randomnas}                  & 124.30 & 1.86 & 0.861 & 0.516 & 57.87  & $-$62.59\% \\
Progressive Search                                    & 127.70 & 1.72 & 0.961 & 0.515 & 55.86  & $-$61.57\% \\
Adaptive Exploration                                  & \textbf{79.12} & 1.72 & 0.980 & 0.531 & 56.56 & $\mathbf{-76.19\%}$ \\
Multi-fidelity~\cite{li2017hyperband}                 & 123.67 & \textbf{2.08} & \textbf{1.00} & 0.512 & 125.88 & $-$62.78\% \\
AdversarialNAS$^\dagger$~\cite{gao2019adversarialnas} & 109.29 & 1.84 & \textbf{1.00} & 0.504 & 124.50$^\dagger$ & $-$67.11\% \\
\bottomrule
\end{tabular}
\end{table*}

All five NAS strategies improve FID over the manual baseline. Even Random
Search reduces FID by 62.59\% in under 58 minutes; principled sampling of
the search space already beats a single manual design point.

Adaptive Exploration gives the largest FID improvement overall ---
$332.28 \to 79.12$ ($76.19\%$) --- in $56.56$ minutes, the same budget
as Random and Progressive Search. It also has the best Recall among
the NAS strategies ($0.531$). The manual baseline scores slightly
higher ($0.552$), but paired with much worse FID and lower Precision
($0.693$), so its ``diversity'' reflects low-fidelity output rather
than genuine coverage.

Multi-fidelity Search attains the highest IS (2.08), just above the manual
baseline (2.00). Thus NAS sharply improves distributional similarity (FID)
while the baseline stays competitive on IS --- the known FID--IS
divergence~\cite{borji2019fidsensitivity}: IS rewards confident,
class-distinct predictions even under poor coverage, whereas FID measures
the full distributional distance and is the more reliable metric. Adaptive
Exploration and Progressive Search score 1.72 on IS, trading it for the FID
gains that are this work's primary criterion.

AdversarialNAS reduces FID by 67.11\% with perfect Precision (1.00); its
124.50-minute search reflects gradient-based optimization of one
architecture and is not directly comparable to the sampling strategies.

Per-strategy training loss curves are omitted here for space; their key
signal --- convergence stability and generator--discriminator balance ---
is already quantified by the Stability and Balance components of the
Enhanced Score (Section~\ref{sec:method:metrics}), and the full curves are
available with the released code.

\subsection{Computational Efficiency}
\label{sec:results:efficiency}

Search time refers to the filtering stage --- training and scoring all
candidates before the best architecture proceeds to 500-epoch full
training.

Among the sampling strategies, Adaptive Exploration, Random, and
Progressive Search all finish within 56--58 minutes, so history-guided
sampling adds no meaningful overhead over uniform sampling. Multi-fidelity
Search takes 125.88 minutes: the long runtime indicates that most candidates
cleared the 0.5 Stage~1 threshold and advanced to later stages, so early
stopping saved little here. This matches prior findings that multi-fidelity methods save most
when many configurations fail early~\cite{jamieson2016successivehalving};
larger or denser search spaces would benefit more.

AdversarialNAS records 124.50 minutes, reflecting ${\sim}300$ epochs of
gradient-based optimization of one architecture --- depth of search rather
than breadth --- and is not directly comparable on search time.

The manual baseline needs no search but fixes a single one of 128 generator
configurations; exhaustively exploring the joint $G$--$D$ space of 4{,}096
pairs is infeasible on a single GPU. NAS recovers a substantially better
configuration in under 60 minutes.

\begin{figure}[t!]
\centering
\includegraphics[width=0.95\columnwidth]{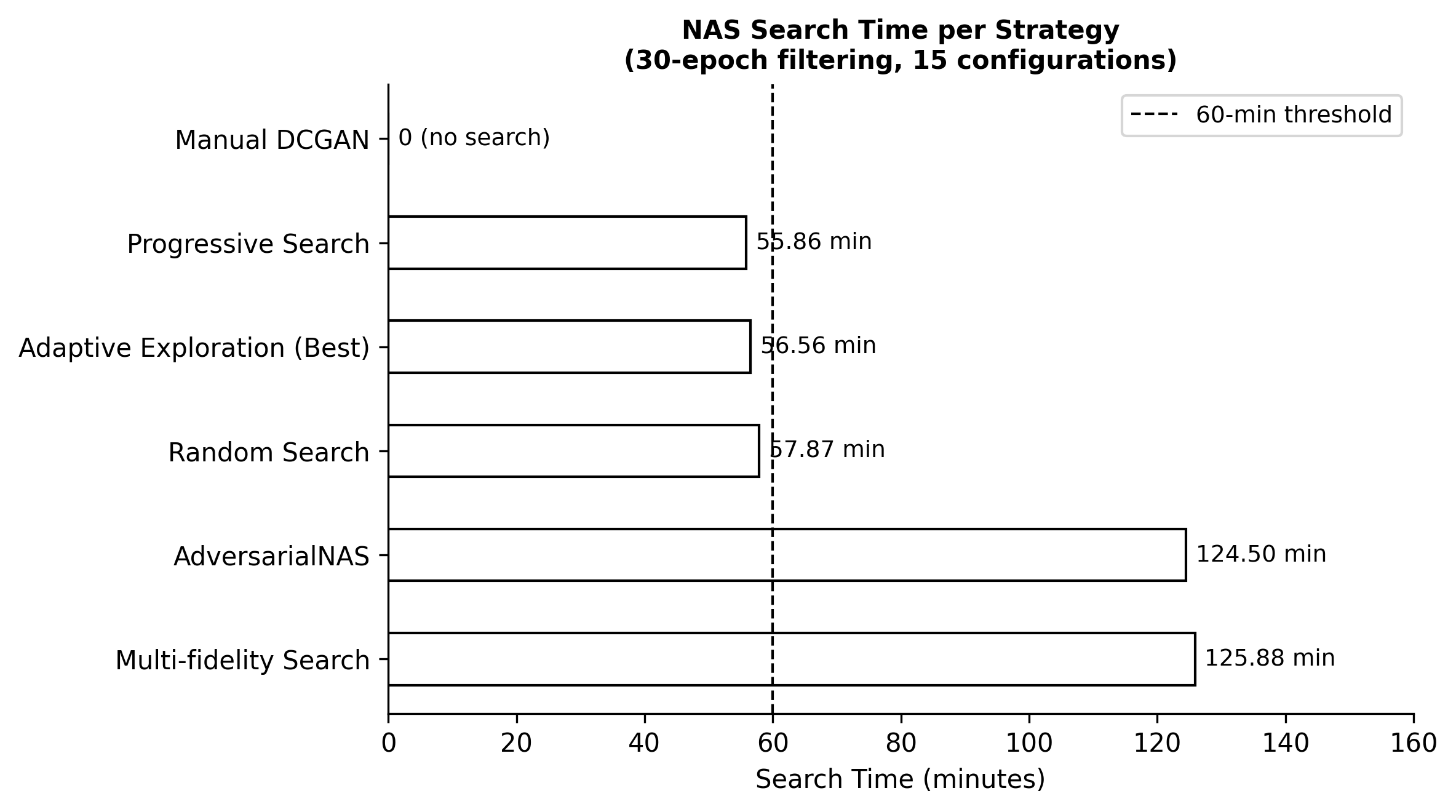}
\caption{Search time (minutes) per NAS strategy
         during the filtering stage.
         Dashed line marks the 60-minute threshold.}
\label{fig:search_time}
\end{figure}

\subsection{Architecture Analysis}
\label{sec:results:arch_analysis}

Table~\ref{tab:full_arch} presents the complete layer-by-layer architecture
of the best NAS-discovered model (Adaptive Exploration), compared against
the manual DCGAN baseline. The Generator $G$ progressively upsamples from
a $128$-dimensional latent vector concatenated with a one-hot class label
through three residual blocks and transposed convolution stages, reaching
a $1\times32\times32$ output image. The Discriminator $D$ mirrors this
structure, downsampling the concatenated image-label input through three
residual blocks to a scalar real/fake score.

We note that the first discriminator block (Down-1) carries no
normalization. This is not a searched choice but a fixed convention we
adopt from standard DCGAN practice~\cite{radford2015dcgan}: the input layer
operates directly on raw pixels, where normalization would standardize away
the low-level intensity statistics the discriminator relies on; all deeper
blocks retain the searched normalization. The searched dimensions thus apply
to layers Down-2 onward, with the raw-pixel layer held fixed.

\begin{table*}[t]
\centering
\caption{Best NAS-discovered architecture (Adaptive Exploration)}
\label{tab:full_arch}
\begin{minipage}[t]{0.49\textwidth}
\centering
\resizebox{\linewidth}{!}{%
\begin{tabular}{llcccl}
\toprule
\multicolumn{6}{c}{\textbf{Generator $G$}\;($d_z{=}128$)}\\
\midrule
\textbf{Block} & \textbf{Operation} & \textbf{Output} & \textbf{Norm} & \textbf{Act.} & \textbf{Add-ons}\\
\midrule
Input   & $z_{128}\oplus y \to \mathbb{R}^{256}$ & $256{\times}1{\times}1$   & -- & --   & --\\
Initial & ConvT $256{\to}512$ & $512{\times}4{\times}4$   & BN & ReLU & --\\
Up-1    & ConvT $512{\to}256$ & $256{\times}8{\times}8$   & BN & ReLU & Drop $0.2$, ResBlock\\
Up-2    & ConvT $256{\to}128$ & $128{\times}16{\times}16$ & BN & ReLU & ResBlock\\
Output  & ConvT $128{\to}1$   & $1{\times}32{\times}32$   & -- & Tanh & --\\
\bottomrule
\end{tabular}}
\end{minipage}
\hfill
\begin{minipage}[t]{0.49\textwidth}
\centering
\resizebox{\linewidth}{!}{%
\begin{tabular}{llcccl}
\toprule
\multicolumn{6}{c}{\textbf{Discriminator $D$}}\\
\midrule
\textbf{Block} & \textbf{Operation} & \textbf{Output} & \textbf{Norm} & \textbf{Act.} & \textbf{Add-ons}\\
\midrule
Input  & $x_{1\times32\times32}\oplus y \to 2$ ch & $2{\times}32{\times}32$ & --   & --     & --\\
Down-1 & Conv $2{\to}64$    & $64{\times}16{\times}16$ & \textit{none} & LReLU & Drop $0.5$, ResBlock\\
Down-2 & Conv $64{\to}128$  & $128{\times}8{\times}8$  & BN & LReLU & Drop $0.5$, ResBlock\\
Down-3 & Conv $128{\to}256$ & $256{\times}4{\times}4$  & BN & LReLU & Drop $0.5$, ResBlock\\
Output & Conv $256{\to}1$   & scalar                   & -- & Sigmoid & --\\
\bottomrule
\end{tabular}}
\end{minipage}

\vspace{2pt}
{\footnotesize\itshape Conv/ConvT: $4{\times}4$, stride 2, pad 1 (Initial \& $D$-output: stride 1, no pad).
ResBlock: $2{\times}$Conv $3{\times}3$ + identity skip.}
\end{table*}

\subsection{Qualitative Results}
\label{sec:results:qualitative}

\begin{figure}[t!]
\centering
\includegraphics[width=0.75\columnwidth]{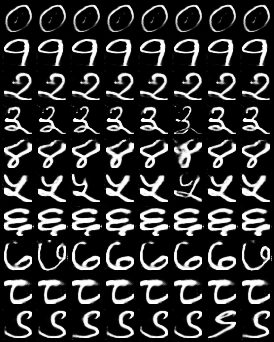}
\caption{
    Generated Devanagari digits from Adaptive Exploration (500-epoch
    checkpoint). One row per digit class (0--9), random samples.
}
\label{fig:adaptive_results}
\end{figure}

Adaptive Exploration's generated digits show clear, legible strokes across
all ten classes. The \textit{shirorekha} (horizontal headline) is
consistently reproduced and inter-class boundaries are well preserved ---
each row is a visually distinct digit, confirming effective conditioning on
$y$. Some within-class variation in stroke thickness (notably digits 2 and
3) is consistent with the Precision of $0.980$, while the Recall of $0.531$
reflects broad coverage with no visible mode collapse.

\subsection{Downstream Classification Utility of GAN-Generated Data}
\label{sec:results:downstream}

To validate the practical utility of NAS-generated digits beyond
distributional metrics, we train a lightweight CNN classifier under
a low-resource setting (250 real samples per class) with and without
GAN augmentation from the best NAS model (Adaptive Exploration).

\begin{table}[H]
\centering
\caption{CNN classification results in low-resource setting
         (250 real samples per class;
          test set: 3{,}000 samples, 300 per class;
          30 epochs, batch size 32).}
\label{tab:cnn_results}
\begin{tabular}{lcc}
\toprule
\textbf{Training Data} & \textbf{Real only} & \textbf{Real + GAN} \\
\midrule
Accuracy    & 91.0\% & \textbf{96.5\%} \\
Precision   & 91.1\% & \textbf{96.6\%} \\
Recall      & 91.0\% & \textbf{96.5\%} \\
F1          & 91.0\% & \textbf{96.5\%} \\
\midrule
$\Delta$ Accuracy & \multicolumn{2}{c}{$\mathbf{+5.6}$ percentage points} \\
\bottomrule
\end{tabular}
\end{table}

Augmenting the 250 real samples per class with GAN-generated digits raises
CNN accuracy from $91.0\%$ to $96.5\%$ ($+5.6$ percentage points). The gain
holds across precision, recall, and F1: augmentation raises overall
classifier quality rather than trading one metric for another. Both
conditions already apply standard image augmentation (random rotation and
translation; Section~\ref{sec:method:cnn}), so this is the benefit of
GAN-generated samples \emph{on top of} conventional augmentation, not a
replacement for it. The classifier is deliberately lightweight to model a
resource-constrained deployment; a higher-capacity classifier would narrow
the absolute headroom, so we read the $+5.6$-point gain as evidence that the
synthetic samples are useful rather than as an upper bound on the effect.

\section{Discussion}
\label{sec:discussion}

\subsection{Why Adaptive Exploration Outperforms All Strategies}
\label{sec:discussion:adaptive}

The 76.19\% FID improvement of Adaptive Exploration over the manual
baseline stems from two factors. First, it evaluates 15 architectures
rather than one, covering a meaningful fraction of the joint
generator--discriminator space (4{,}096 pairs) that manual design cannot.
Second, its history-guided mutation biases later candidates toward regions
that scored well on the Enhanced Score, avoiding unstable or
mode-collapsing configurations early.

\subsection{The Quality-Diversity Trade-off}
\label{sec:discussion:tradeoff}

The four metrics trace a quality-diversity spectrum rather than a
single winner~\cite{kynkaanniemi2019precision}.

On FID, Adaptive Exploration leads clearly ($79.12$ vs.\ $62$--$67\%$
gains for the others), suggesting history-guided sampling is
especially effective at closing the overall distributional gap.

On IS, Multi-fidelity Search edges out even the manual baseline
($2.08$ vs.\ $2.00$). This reflects how IS is scored: locally sharp,
confident predictions from the Inception classifier can occur even
without good distributional coverage, so a high-FID model can still
post a decent IS. NAS methods like Adaptive Exploration trade some of
that sharpness for better FID, which we treat as the more reliable
metric~\cite{borji2019fidsensitivity}.

On Precision, AdversarialNAS and Multi-fidelity both hit $1.00$ ---
their gradient-based and multi-stage filtering actively suppress
low-fidelity samples, at the cost of Recall.

On Recall, Adaptive Exploration leads among NAS methods ($0.531$);
only the manual baseline is higher ($0.552$), for the reason noted
above. AdversarialNAS is lowest ($0.504$), consistent with its
narrower, high-confidence optimization.

In practice: use Adaptive Exploration for overall quality and
coverage, AdversarialNAS or Multi-fidelity when low-fidelity samples
are costly (e.g., augmentation pipelines), and Multi-fidelity when IS
is the target metric.

\subsection{The Role of the Enhanced Score}
\label{sec:discussion:enhanced}

A key design choice in NepScript Genesis is using the Enhanced Score
(Equation~\ref{eq:enhanced_score}) rather than a proxy FID for architecture
filtering. The Enhanced Score is computable after only 30 epochs and
captures training health signals (Stability, Balance) that predict final
model quality without requiring the full 500-epoch run. The inclusion of
Nepali Quality ($Q_{\text{Nepali}}$) as a domain-specific component
ensures that architectures are rewarded for producing legible Devanagari
strokes --- a criterion that generic metrics like IS would not capture.

\subsection{Limitations}
\begin{itemize}[leftmargin=*, itemsep=2pt]
    \item \textbf{Search space scope:} Our 128-configuration search space
    covers only five dimensions. Deeper networks, attention mechanisms, and
    alternative loss functions (e.g., Wasserstein~\cite{arjovsky2017wgan},
    spectral normalization~\cite{miyato2018spectralnorm}) are excluded.

    \item \textbf{Single-seed results:} Due to compute constraints, each
    architecture is evaluated with a single training run. Multi-seed
    variance estimation~\cite{pineau2021reproducibility} would strengthen
    the reliability of the reported numbers.

    \item \textbf{Script specificity:} Results are on Devanagari digits;
    generalization to the full Devanagari character set or other Brahmic
    scripts requires additional validation~\cite{bhunia2019lowresource}.

    \item \textbf{Evaluation metric sensitivity:} FID is known to be
    sensitive to the number of samples and the choice of Inception
    network~\cite{borji2019fidsensitivity}. All FID values in this paper
    are computed on $20{,}000$ samples to minimize this sensitivity.
\end{itemize}

\section{Conclusion}
\label{sec:conclusion}

We presented NepScript Genesis, a NAS framework for automated GAN
architecture discovery for conditional Devanagari handwritten digit
synthesis. All five NAS strategies outperform the manually designed
DCGAN baseline: within a fixed compute budget, automated search
consistently finds better configurations than a single hand-crafted
design. Adaptive
Exploration achieves the best overall result --- FID $= 79.12$, a
76.19\% improvement over the manual baseline (FID $= 332.28$) ---
while matching the search time of the simplest strategies (56.56 min).
AdversarialNAS delivers perfect Precision
(1.00) through gradient-based joint optimization of $G$ and $D$, and
Multi-fidelity Search achieves the highest IS
(2.08). The two-stage evaluation pipeline --- 30-epoch Enhanced Score
filtering followed by 500-epoch full training --- provides an efficient
and domain-aware mechanism for architecture selection, with the Nepali
Quality component ($Q_{\text{Nepali}}$) ensuring that Devanagari stroke
characteristics are explicitly rewarded during search. In a downstream low-resource evaluation,
augmenting 250 real training samples per class with
GAN-generated digits from Adaptive Exploration improves CNN
classification accuracy by $+5.5$ percentage points
($91.0\%\to96.5\%$), confirming that NAS-optimized synthesis
produces digits of sufficient quality to benefit practical
recognition pipelines when real data is scarce

\bibliography{references}

\end{document}